\documentclass[10pt,twocolumn,letterpaper]{article}

\usepackage[pagenumbers]{cvpr} 

\usepackage[dvipsnames]{xcolor}

\definecolor{cvprblue}{rgb}{0.21,0.49,0.74}
\usepackage[pagebackref,breaklinks,colorlinks,citecolor=cvprblue]{hyperref}
\usepackage{multirow}
\usepackage{booktabs}
\usepackage{graphicx}

\usepackage{graphicx}
\usepackage[utf8]{inputenc} 
\usepackage[T1]{fontenc}    
\usepackage{url}            
\usepackage{booktabs}       
\usepackage{amsfonts}       
\usepackage{nicefrac}       
\usepackage{microtype}      
\usepackage{epsfig}
\usepackage{float}
\usepackage{multicol}
\usepackage{multirow}
\usepackage{amssymb}
\usepackage{amsmath}
\usepackage{wrapfig}
\usepackage{xspace}
\usepackage{color}
\usepackage{colortbl}
\usepackage[dvipsnames, svgnames, x11names, table]{}
\usepackage{xcolor}
\usepackage[misc]{ifsym}
\usepackage{makecell}
\usepackage{soul}
\usepackage{bm}
\usepackage{wrapfig}
\usepackage{subcaption, overpic, textpos}

\definecolor{method_orange}{RGB}{222, 131, 68}
\definecolor{method_green}{RGB}{126, 171, 85}
\definecolor{method_blue}{RGB}{106, 153, 208}

\def\paperID{3453} 
\def\confName{CVPR}
\def\confYear{2024}
\newcommand{\name}{$\mathtt{FaceX}$}

\title{A Generalist \textbf{$\mathtt{FaceX}$} via Learning Unified Facial Representation}

\author{Yue Han\textsuperscript{1$\ast$} 
\quad Jiangning Zhang\textsuperscript{2$\ast$} 
\quad Junwei Zhu\textsuperscript{2} 
\quad Xiangtai Li\textsuperscript{3} 
\quad Yanhao Ge\textsuperscript{4}  \\
\quad Wei Li\textsuperscript{4} 
\quad Chengjie Wang\textsuperscript{2} 
\quad Yong Liu\textsuperscript{1$\dagger$} 
\quad Xiaoming Liu\textsuperscript{5}
\quad Ying Tai\textsuperscript{6}\\
\normalsize \textsuperscript{1}{APRIL Lab, Zhejiang University} \quad \textsuperscript{2}{Youtu Lab, Tencent} \quad \textsuperscript{3}{Nanyang Technological University} \\ 
\normalsize \textsuperscript{4}{VIVO} \quad \textsuperscript{5}{Michigan State University} \quad \textsuperscript{6}{Nanjing University} \\
{\normalsize *: equal contribution \quad $\dagger$: corresponding author.} \\
{\normalsize Project Page: \url{https://diffusion-facex.github.io}}
\vspace{-1.5em}
}

\begin{document}

\twocolumn[{%
\renewcommand\twocolumn[1][]{#1}%
\maketitle
\begin{center}
    \centering
    \captionsetup{type=figure, skip=0pt} 
    \includegraphics[width=\textwidth]{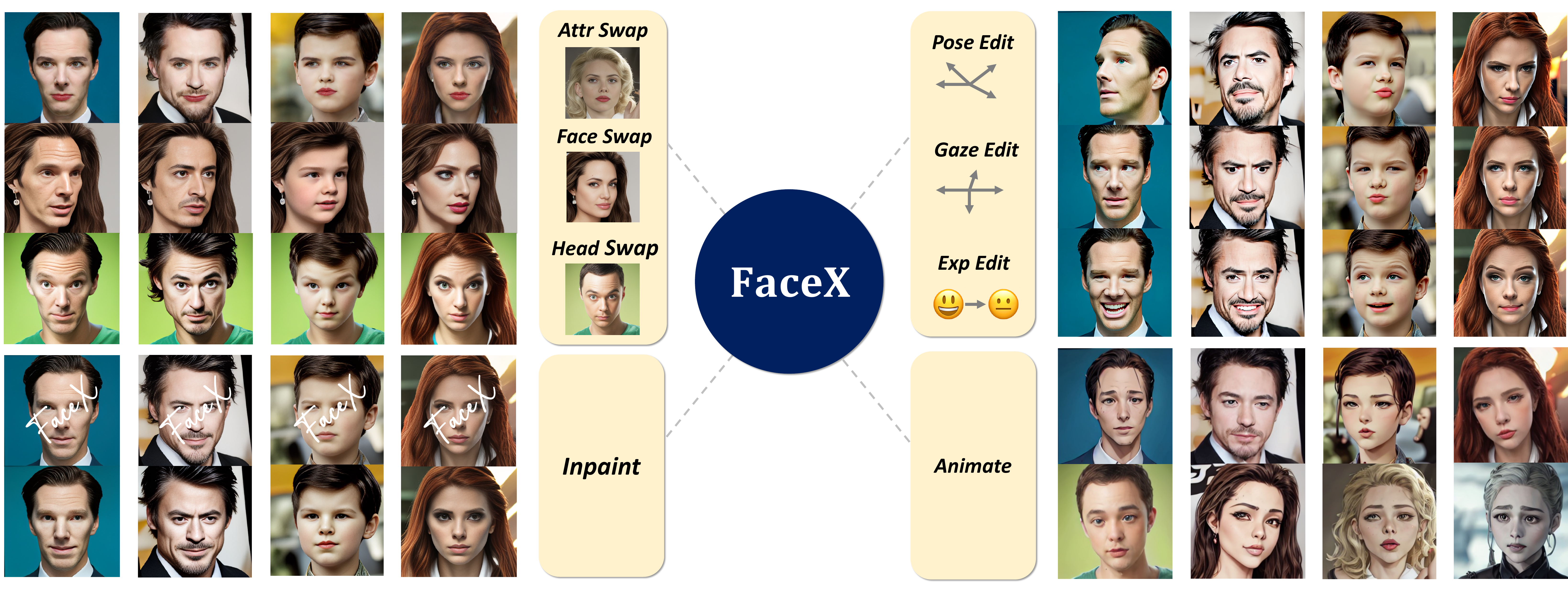} 
    \captionof{figure}{
     \textbf{Facial generalist \name~is capable of handling diverse facial tasks}, ranging from popular face/head swapping and motion-aware face reenactment/animation to semantic-aware attribute editing/inpainting, by one unified model, simultaneously achieving competitive performance that significantly advances the research of general facial models.
    }
\label{fig:teaser}
\end{center}%
}]

\begin{abstract}
This work presents \name~framework, a novel facial generalist model capable of handling diverse facial tasks simultaneously.
To achieve this goal, we initially formulate a unified facial representation for a broad spectrum of facial editing tasks, which macroscopically decomposes a face into fundamental identity, intra-personal variation, and environmental factors. 
Based on this, we introduce Facial Omni-Representation Decomposing (FORD) for seamless manipulation of various facial components, microscopically decomposing the core aspects of most facial editing tasks.
Furthermore, by leveraging the prior of a pretrained StableDiffusion (SD) to enhance generation quality and accelerate training, we design Facial Omni-Representation Steering (FORS) to first assemble unified facial representations and then effectively steer the SD-aware generation process by the efficient Facial Representation Controller (FRC). 
Our versatile~\name~achieves competitive performance compared to elaborate task-specific models on popular facial editing tasks. 
Full codes and models are available at \url{https://github.com/diffusion-facex/FaceX}.

\end{abstract}

\vspace{-3mm}    
\section{Introduction}
\label{sec:introduction}

\begin{figure*}[htp]
    \includegraphics[width=\textwidth]{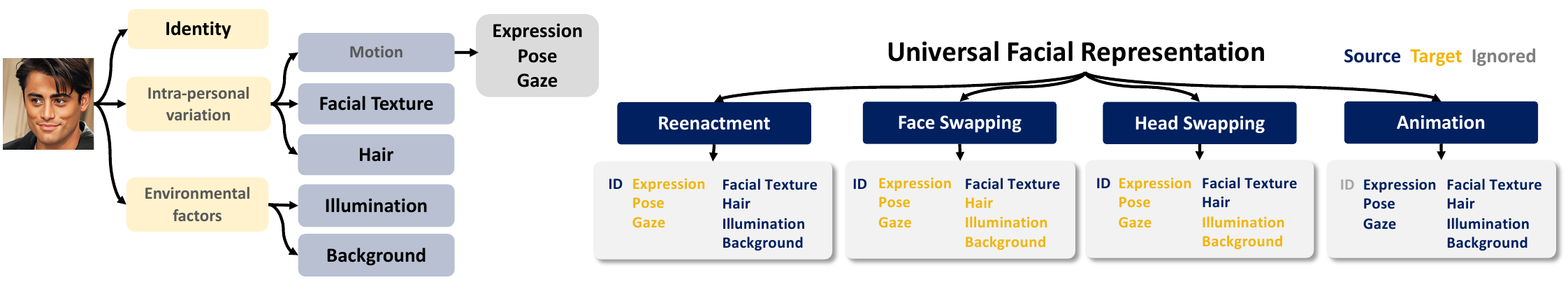}
    \vspace{-6mm}
    \caption{\textbf{Left}: Proposed facial omni-representation equation that divides one face into a combination of different fine-grained attributes. \textbf{Right}: The attributes of the generated images under different tasks correspond to the decomposition of source and target facial attributes.
    Here, we analyze four representative facial tasks. 
    For details of other facial tasks, please refer to our supplementary materials.
    } 
    \label{fig:intro}
\end{figure*}


Facial editing encompasses both low-level tasks, \eg, facial inpainting~\cite{zhang2021inpainting} and domain stylization~\cite{gal2022stylegan}, and high-level tasks, \eg, region-aware face/head/attribute swapping~\cite{wang2021hififace,liu2023e4s,shu2022heser,luo2022styleface,nirkin2022fsganv2}, motion-aware pose/gaze/expression control~\cite{zhu2022hifihead,zhang2020freenet,xu2023emotiontalk}.
Above tasks have extensive applications in various domains, including entertainment, social media, and security.
The primary challenge in facial editing is to modify distinct attributes while preserving identity and unaffected attributes consistently.
Notably, there's also a need for in-the-wild generalization to ensure practical applicability.

Previous GAN-based methods leverage the disentangled latent space of StyleGAN~\cite{karras2020stylegan2}, enabling attribute manipulation by navigating within the latent space along suitable directions.
Thanks to the powerful generative capabilities of Diffusion Models (DM), recent works have embraced this technique for enhancing the quality of facial generation in various editing tasks. 
However, disentangling and controlling facial attributes using DM in a zero-shot manner remains an unresolved issue. 
%
For example, Face$0$~\cite{valevski2023face0} enables one-shot identity insertion but struggles with attribute disentanglement. 
DiffusionRig~\cite{ding2023diffusionrig} achieves pose/expression control by physical DECA~\cite{feng2021deca}, but requires a time-consuming fine-tuning procedure for identity generalization. 
DiffTalk~\cite{shen2023difftalk} relies on landmark-guided inpainting to keep other parts intact. 
Recent DiffSwap~\cite{zhao2023diffswap} uses identity feature along with an identity loss to maintain identity and employs DDIM~\cite{song2020ddim} inversion to preserve other parts.
%
%
%
%
The above methods are designed with elaborate modules tailored for specific tasks, rendering them {\textit{challenging to generalize across different tasks}}, thereby limiting their versatility and increasing the R\&D cost in practical applications. 
In contrast, universal models, with higher practical value, have garnered significant success in the fields of NLP~\cite{brown2020language,ouyang2022training} and segmentation~\cite{kirillov2023segment}. 
However, \textit{the absence of a universal facial editing model persists due to the diverse nature of facial tasks}.

%
To address this issue, for the first time, we present a generalist facial editing model, termed~\name.
Our method handles \textit{extensive} facial editing tasks with a \textit{unified} model (see Fig.~\ref{fig:teaser}), while maintaining the ability to disentangle and edit various attributes when generating high-quality images. 
Specifically, there are two significant designs in our~\name:

\noindent $1)$ \textbf{Facial Omni-representation Decomposing}: 
We establish a coherent facial representation for a wide range of facial editing tasks, inspired by probabilistic LDA~\cite{ioffe2006plda, prince2011plda}. 
Our solution introduces a unified facial representation equation to macroscopically decompose a face into three factors:
\begin{equation}
    \begin{aligned}
        \mathtt{X} =  \mathcal{G}(\alpha, \beta, \gamma),
    \end{aligned}
    \label{eq:eq1}
\end{equation}
where identity $\alpha$, intra-personal variation $\beta$, and environmental factors $\gamma$ are fundamental attributes that characterize a face $\mathtt{X}$. 
$\mathcal{G}$ indicates a powerful generative model.
%
%
Furthermore, we assume that the intra-personal variation can be decomposed into motion, facial texture, and hair, while environmental factors corresponde to illumination and background. 
As shown in~\cref{fig:intro},~\name~enables clear formula-level task decomposition, easy manipulation, and quick adaptation to various facial editing tasks, making a versatile and efficient solution possible. 
More specifically, we adopt pretrained face recognition model~\cite{deng2019arcface} to achieve identity feature, pretrained D$3$DFR model~\cite{deng2019accurate} to obtain $3$D coefficients for motion variations, and a vision image encoder (\eg, DINOV$2$~\cite{oquab2023dinov2} or CLIP~\cite{radford2021clip}) to model the textures of facial, hair and environmental comprehensively. 
Leveraging our disentangled omni-representation, we can manipulate different features for diverse  editing tasks, \cf, \cref{sec:fors}.

\noindent $2)$ \textbf{Steering and Controlling Omni-representation in DM}: 
With the proposed universal facial representation, a core challenge is how to {extract} and {utilize} it to control the generation process of DM. 
Specifically, we utilize the prior of a pretrained StableDiffusion (SD) to enhance generation quality and accelerate training. 
Existing methods augmenting conditional control in SD employ different fine-tuning approaches:
\textit{i)} The intuitive approach concatenates input and noise latent, and fine-tunes the \textit{entire} U-net, which incurs significant training costs. 
\textit{ii)} ControlNet~\cite{zhang2023controlnet} and T2I-Adapter~\cite{mou2023t2i-adapter} fine-tune \textit{additional} encoders while fixing the U-net. 
However, they are only suitable for localized control, lacking low-level texture control. 
\textit{iii)} Text-guided control effectively alters texture, but mapping facial representation to the CLIP text domain with a fixed U-net~\cite{ronneberger2015unet} \textit{fails} at texture reconstruction. 
Inspired by the gated self-attention in GLIGEN~\cite{li2023gligen} with grounding conditions, we propose a powerful Facial Omni-Representation Steering module (\cref{sec:fors}) to aggregate task-specific rich information from the input facial images, and then design an efficient and effective Facial Representation Controller (\cref{sec:frc}) to enable Style Diffusion to support fine-grained facial representation modulation.


%
Overall, our contribution can be summarized as follows:
\begin{itemize}
  \item To our best knowledge, the proposed~\name~is the first generalist facial editing model that seamlessly addresses a variety of facial tasks through a single model. 
  \item We propose a unified facial representation to macroscopically formulate facial compositions, and further design a Facial Omni-Representation Decomposing (FORD) module to microscopically decompose the core aspects of most facial editing tasks to easily manipulate various facial details, including ID, texture, motion, attribute, \etc. 
  \item We introduce the Facial Omni-Representation Steering (FORS) to first assemble unified facial representations and then effectively steer SD-aware generation process by the efficient Facial Representation Controller (FRC).
  \item Extensive experiments on eight tasks validate the unity, efficiency, and efficacy of our~\name. Ablation studies affirm the necessity and effectiveness of each module.
\end{itemize}

\section{Related Works}
\label{sec:related_works}

\noindent
\textbf{Diffusion Models} have made significant progress in image generation, demonstrating exceptional sample quality~\cite{ho2020denoising}. 
%
Employing a denoising process through the U-Net structure, these models iteratively refine Gaussian noise to generate clean data.
%
However, the quadratic growth in memory and computational demands, primarily due to self-attention layers in the U-Net, is a challenge escalated with increasing input resolution.
Recent advancements emphasize speeding up the training and sampling of DMs. 
Latent DMs (LDMs)~\cite{rombach2022high} are trained in a latent embedding space instead of the pixel space. 
Additionally, LDMs introduce cross-attention among conditional input feature maps at multiple resolutions in the U-Net, effectively guiding denoising.

\noindent \textbf{Face Editing} 
encompasses both low- and high-level tasks~\cite{zeng2020realistic,zhang2021inpainting,zhang2020apb2face,zhang2021real,li2021faceinpainter,gal2022stylegan,liu2021blendgan,chen2020simswap,wang2021hififace,liu2023e4s,shu2022heser,xu2022region,xu2022designing,luo2022styleface,nirkin2022fsganv2,xu2022high,zhu2022hifihead,zhang2020freenet,xu2023emotiontalk,zhang2023rethinking}.
DifFace~\cite{yue2022difface} retrains the DM from scratch on pre-collected data for face restoration.
Face$0$~\cite{valevski2023face0} facilitates one-shot identity insertion and text-based facial attribute editing. 
DiffuionRig~\cite{ding2023diffusionrig} achieves pose and expression control via physical buffers of DECA~\cite{feng2021deca} but requires finetuning for identity generalization.
DiffTalk~\cite{shen2023difftalk} relies on landmarks and inpainting for talking face generation when the mouth region is driven by audio.
DiffSwap~\cite{zhao2023diffswap} leverages landmarks to control expression and pose, uses face  ID features as conditions, and relies on a single denoising step loss to maintain identity. 

Existing facial editing tasks encounter common challenges, involving disentangling and editing different attributes, preserving identity or other non-edited attributes during editing, and facilitating generalization for real-world applications.
Therefore, instead of adopting the conventional \textit{single-model-single-task} approach, we comprehensively model facial representations and establish a unified editing framework, supporting \textit{single-model-multi-task} scenarios.

\noindent \textbf{Condition-guided Controllable SD} 
The incorporation of conditions can be primarily divided into four categories:
$1$) Concatenating the control conditions at the input and fully fine-tuning the U-Net is suitable for localized conditions but significantly increases the training cost, \eg HumanSD~\cite{ju2023humansd} and Composer~\cite{huang2023composer}.
$2$) Projecting and adding conditions to the timestep embedding or concatenating them with CLIP~\cite{radford2021clip} word embeddings, used as context input for cross-attention layers, is effective for global conditions such as intensity, color, and style. 
However, fine-tuning the entire U-Net with text-condition pairs (\eg, Composer~\cite{huang2023composer}), incurs high training cost, while fixing U-Net requires optimization for each condition.
$3$) Fine-tuning additional encoders while fixing U-Net is suitable for localized control but not for low-level texture control (\eg, ControlNet~\cite{zhang2023controlnet}, T2I-Adapter~\cite{mou2023t2i-adapter}, and LayoutDiffusion~\cite{zheng2023layoutdiffusion}).
$4$) Introducing extra attention layers in U-Net to incorporate conditions, \eg, GLIGEN~\cite{li2023gligen}.
In this paper, we adopt a method akin to GLIGEN for incorporating unified facial representation, empirically demonstrating its efficiency and effectiveness.
\section{Methods} \label{sec:methods}

\subsection{Preliminary Diffusion Models} \label{sec:diffusion}
Denoising Diffusion Probabilistic Models (DDPMs) are a class of generative models, which recovers high-quality images from Gaussian noise (\ie, denoising process) by learning a reverse Markov Chain (\ie, diffusion process): $\boldsymbol{x}_t$$\sim$$\mathcal{N}\left(\sqrt{\alpha_t} \boldsymbol{x}_{t-1},\left(1-\alpha_t\right) \boldsymbol{I}\right)$, where $\boldsymbol{x}_t$ is the random variable at $t$-th timestep and $\alpha_t$ is the predefined coefficient. 
In practice, $\boldsymbol{x}_t=\sqrt{\bar{\alpha}_t} \boldsymbol{x}_0+\sqrt{1-\bar{\alpha}_t} \boldsymbol{\epsilon}$ is used as approximation to facilitate efficient training, where $\bar{\alpha}_t=\prod_{s=1}^t \alpha_s$ and $\boldsymbol{\epsilon}$$\sim$$\mathcal{N}(\mathbf{0}, \boldsymbol{I})$. 
By minimizing the ELBO of the diffusion process, the training objective is simplified to $\mathbb{E}_{\boldsymbol{x}_0, \boldsymbol{\epsilon}, t}\left[\left\|\boldsymbol{\epsilon}-\boldsymbol{\epsilon}_\theta\left(\boldsymbol{x}_t, t\right)\right\|_2^2\right]$. 
In the inference, U-Net-based denoising autoencoder $\boldsymbol{\epsilon}_\theta\left(x_t, t\right)$ is predicted step by step to obtain the final $\boldsymbol{x}_0$. 
As  naive DDPMs are computationally costly,
Latent Diffusion Model (LDM)~\cite{ldm} proposes to train the model in the latent space $\boldsymbol{z}$ compressed by VQGAN~\cite{vqgan}, whose basic paradigm is also adopted in this paper.

\begin{figure*}[tp]
    \centering
    \includegraphics[width=1.0\linewidth]{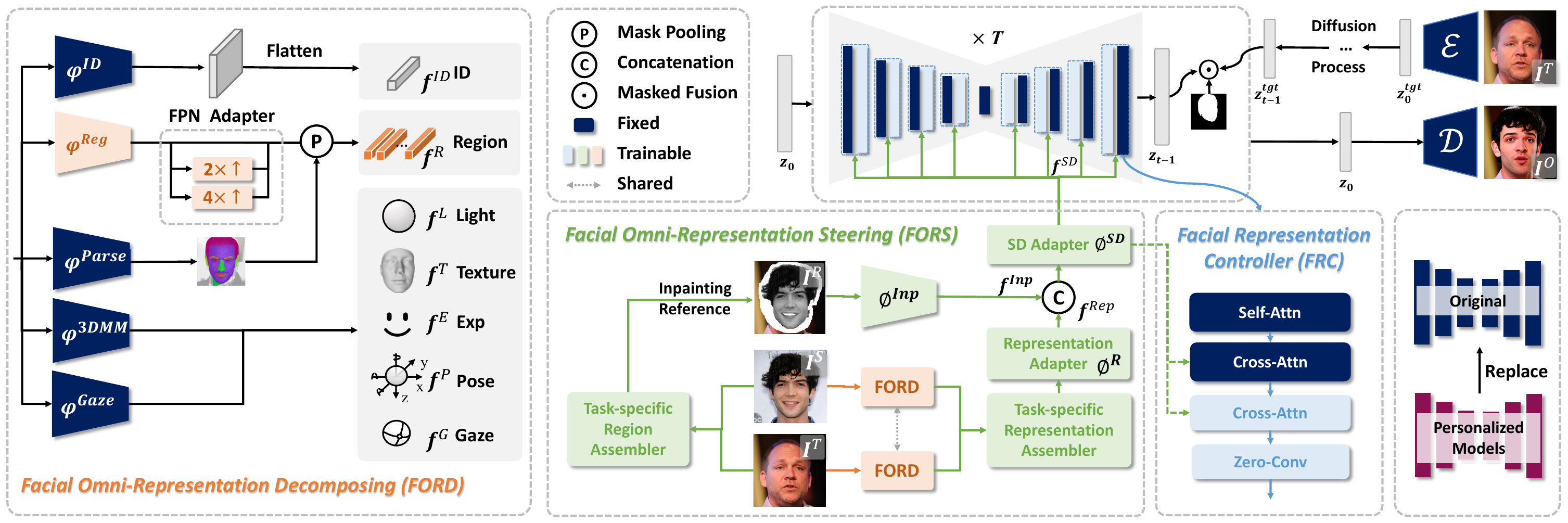}
    \vspace{-6mm}
    \caption{\textbf{Overview of the \name~framework}, which consists of: 
    \textbf{\textit{1)}} \textcolor{method_orange}{\textbf{Facial Omni-Representation Decomposing} (FORD)}  $\bm{\varphi}=\{\bm{\varphi}^{ID}, \bm{\varphi}^{Reg}, \bm{\varphi}^{Parse}, \bm{\varphi}^{3DMM}, \bm{\varphi}^{Gaze}\}$ decomposes facial component representations, \ie, $\bm{f}^{ID}$, $\bm{f}^{R}$, $\bm{f}^{L}$, $\bm{f}^{T}$, $\bm{f}^{E}$, $\bm{f}^{P}$, and $\bm{f}^{G}$. 
    \textbf{\textit{2)}} \textcolor{method_green}{\textbf{Facial Omni-Representation Steering} (FORS)} $\bm{\phi}$ contains a Task-specific Representation Assembler to assemble various attributes extracted from source image $\bm{I}^{S}$ and target image $\bm{I}^{T}$, which pass through a Representation Adapter $\bm{\phi}^{R}$ to yield $\bm{f}^{Rep}$; 
    and a Task-specific Region Assembler to assemble different regions to obtain the inpainting reference image $\bm{I}^{R}$, which is then processed by an image encoder $\bm{\phi}^{Inp}$ to obtain $\bm{f}^{Inp}$. After concatenation with $\bm{f}^{Rep}$, it is processed by the SD Adapter $\bm{\phi}^{SD}$ to obtain the conditional representation $\bm{f}^{SD}$ that is fed into the conditional denoising U-Net $\boldsymbol{\epsilon}_\theta$. 
    \textcolor{method_blue}{\textbf{\textit{3)}} \textbf{Facial Representation Controller} (FRC)}, given the basic concatenation of fixed self-/cross-attention operations, we add  one extra cross-attention layer. 
    Under the control of $\bm{f}^{SD}$, it enables generating task-specific output images $\bm{I}^{O}$. 
    Notably, due to the plug-and-play nature of FRC, representations can be seamlessly integrated by cross-attention layers, allowing the diffusion model to be substituted with \textit{any} other personalized models from the community.
    }
    \label{fig:facex}
    \vspace{-1em}
\end{figure*}

\subsection{Facial Omni-Representation Decomposing} \label{sec:ford}
Based on the unified facial representation~\cref{eq:eq1}, we apply it to actual modeling, \ie, we extract different facial components with various pre-trained models. 
As shown on the left side of~\cref{fig:facex}, the unified facial representation include:

\noindent
\textbf{Identity Features.}~We use a face recognition model $\bm{\varphi}^{ID}$~\cite{deng2019arcface} to extract discriminative identity features.
Unlike prior works that select the highly discriminative features of the last layer, we select the uncompressed feature map of the previous layer, which is flattened as the identity embedding $\bm{f}^{ID}$. 
We believe this manner offers richer facial spatial information, while balancing discriminative and generative capabilities.

\noindent
\textbf{Region Features.} In \cref{fig:intro}-Left, the region features include \textit{facial texture, hair, and background}. 
In practical modeling, we further divide facial texture into smaller regions for representation, including \textit{eyebrows, eyes, nose, lips, ears, and skin}.
To align with SD text space, CLIP ViT~\cite{vit,radford2021clip} is used as the encoder $\bm{\varphi}^{Region}$, instead of the commonly used PSP~\cite{psp} in prior works. 
However, compared to the hierarchical structure of PSP, the uniform resolution of ViT limits the spatial information granularity. To address this issue, we employ a \textit{learnable FPN Adapter} to recover the spatial relationships at a higher resolution.
The face parsing model~\cite{to_be_add} $\bm{\varphi}^{Parse}$ is used to obtain regional masks. The region features are extracted via mask pooling. 
Besides CLIP ViT, we also ablate by using ViT from different models in \cref{se:ablation}, finding that pretrained weights and whether to fine-tune significantly impact convergence speed and generated image quality.

\noindent
\textbf{Motion Descriptor.} $3$D pose/expression embedding coefficients $\bm{f}^{P}$/$\bm{f}^{E}$ extracted by the pretrained D$3$DFR model~\cite{deng2019accurate} $\bm{\varphi}^{3DMM}$ and additional gaze embedding $\bm{f}^{G}$ extracted by work~\cite{to_be_add} $\bm{\varphi}^{Gaze}$ form a complete motion descriptor.
Additionally, the disentangled facial texture $\bm{f}^{T}$ and lighting $\bm{f}^{L}$ are used to 
work together with the skin region features to enhance the facial generation quality.

\subsection{Facial Omni-Representation Steering} \label{sec:fors}

The disentangled facial representation can be flexibly recombined for various facial editing tasks, as illustrated in \cref{fig:teaser}. We propose three components to reassemble and fuse features to steer the task-specific generation process. 

\noindent
\textbf{Task-specific Representation Assembler} reassembles the representations of source and target images at the feature level, obtaining the reassembled features $\bm{f}^{Rep}$ via a Representation Adapter $\bm{\phi}^{R}$, which consists of linear layers for each representation to transform the feature dimension for further concatenation.
Complex facial editing tasks, including reenactment, face and head swapping are used as examples here. For all three tasks, the identity features and motion descriptors come from the source and target image respectively. The combination of region features differs for each task, which is detailed in \cref{sec:frc}.
 
Although mask pooling of region features makes appearance editing easier, it results in loss of structural information, leading to increased training difficulty and lack of detail in the generated results. 
To tackle this issue, prior works commonly use masks as structure guidance~\cite{pairdiffusion,sean}. 
However, mask-based structure guidance only supports aligned attribute swapping and struggles to handle motion transformation.
For instance, when swapping a front-facing head onto a side profile, the mask also needs to rotate accordingly. Otherwise, the strong structural constraints will lead to a result where the front-facing face is forcibly squeezed into the side profile. 
HS-diffusion~\cite{HS-diffusion} attempts to address these motion-caused structural changes by training an additional mask converter, but the outcomes are not satisfactory.

\noindent
\textbf{Task-specific Region Assembler} is introduced to tackle this problem. Different regions are assembled at the image level to obtain the region-swapped image $\bm{I}^{R}$, which acts as the inpainting reference for the model. $\bm{I}^{R}$ differs for each task, which is detailed in \cref{sec:frc}.
The inpainting reference $\bm{I}^{R}$ goes through an image encoder $\bm{\phi}^{Inp}$ and obtains the image representation $\bm{f}^{Inp}$. Instead of imposing strong structural constraints through masks, introducing the inpainting reference provides structural clues for the model and meanwhile encourages reasonable imagination. 
Furthermore, this approach introduces additional rich and detailed local structural information, such as hair texture.

\noindent
\textbf{SD Adapter} $\bm{\phi}^{SD}$ adapts the concatenated facial representation to obtain $\bm{f}^{SD}$, effectively steering subsequent SD-aware generation process.

\noindent
\textbf{Diverse and Mixture Editing} is realized by our single model, allowing modifications like glasses, beards, shapes, hairstyles, inpainting, or even their combinations. 
This enhances the interactivity of editing, facilitated by the intuitive image-level region assembler.
To our best knowledge,~\name~stands out as the pioneering work achieving cross-task mixture editing, surpassing the capabilities of existing task-specific methods. 
We hope it serves as a \textit{seed} with potential to \textit{inspire novel and intriguing applications in the future}. 

\subsection{Facial Representation Controller} \label{sec:frc}
For conditional generative models, a core challenge is how to effectively and efficiently use the rich facial representation $\bm{f}^{SD}$ to guide the generation process of the target image $\bm{I}^{O}$. 
Here, we utilize the prior of a pretrained StableDiffusion (SD)~\cite{rombach2022high} to accelerate training and enhance generation quality. 
Unlike recent efficient finetuning schemes~\cite{to_be_add}, we propose a Facial Representation Controller (FRC) module to extend the basic Transformer block in LDM~\cite{ldm}. 
Specifically, the original Transformer block of LDM consists of two attention layers: one self-attention over the visual tokens $\boldsymbol{v}$, followed by cross-attention from context tokens $\boldsymbol{f}^{SD}$. By considering the residual connection, the two layers can be written as: 
\begin{equation}
    \begin{aligned}
        & \boldsymbol{v}=\boldsymbol{v}+\operatorname{SelfAttn_{fix}}(\boldsymbol{v}) \\
        & \boldsymbol{v}=\boldsymbol{v}+\operatorname{CrossAttn_{fix}}\left(\boldsymbol{v}, \boldsymbol{f}^{SD}\right),
    \end{aligned}
    \label{eq:eq2}
\end{equation}
when $\bm{f}^{SD}$ is used as a condition, we empirically find that using only the above two frozen layers can capture coarse identity and motion, but the reconstructed texture detail is very poor, \cf, qualitative results in \cref{fig:visual_abla}-right.
We hypothesize that the reason is that the SD text space is not a continuous, dense facial semantic latent space like StyleGAN, making it challenging to map facial representations to the text space. However, finetuning the entire SD to adapt to the facial domain is computationally expensive, and we want to minimize the loss of SD prior as much as possible.
Therefore, instead of finetuning the original cross-attention layer, we choose to add a new cross-attention layer after the existing one. By only fine-tuning the newly added cross-attention layer, we enable the network to learn to accept facial representations for modulating the intermediate features in the U-net. 
Additionally, we add a zero convolution layer after the newly added cross-attention layer. This way, the starting point of training is equivalent to the original U-net.
\begin{equation}
    \begin{aligned}
        \boldsymbol{v}=\boldsymbol{v}+\operatorname{ZeroConv}\left(\operatorname{CrossAttn_{ft}}\left(\boldsymbol{v}, \boldsymbol{f}^{SD}\right)\right).
    \end{aligned}
    \label{eq:eq3}
\end{equation}
Compared to finetuning the entire SD, this approach is more efficient and effective. 
Moreover, owing to the plug-and-play design, our generalist facial editing model supports loading the personalized models of SD from the community, which can be easily extended to other tasks such as animation.

\begin{figure}[t]
    \centering
    \includegraphics[width=1.0\linewidth]{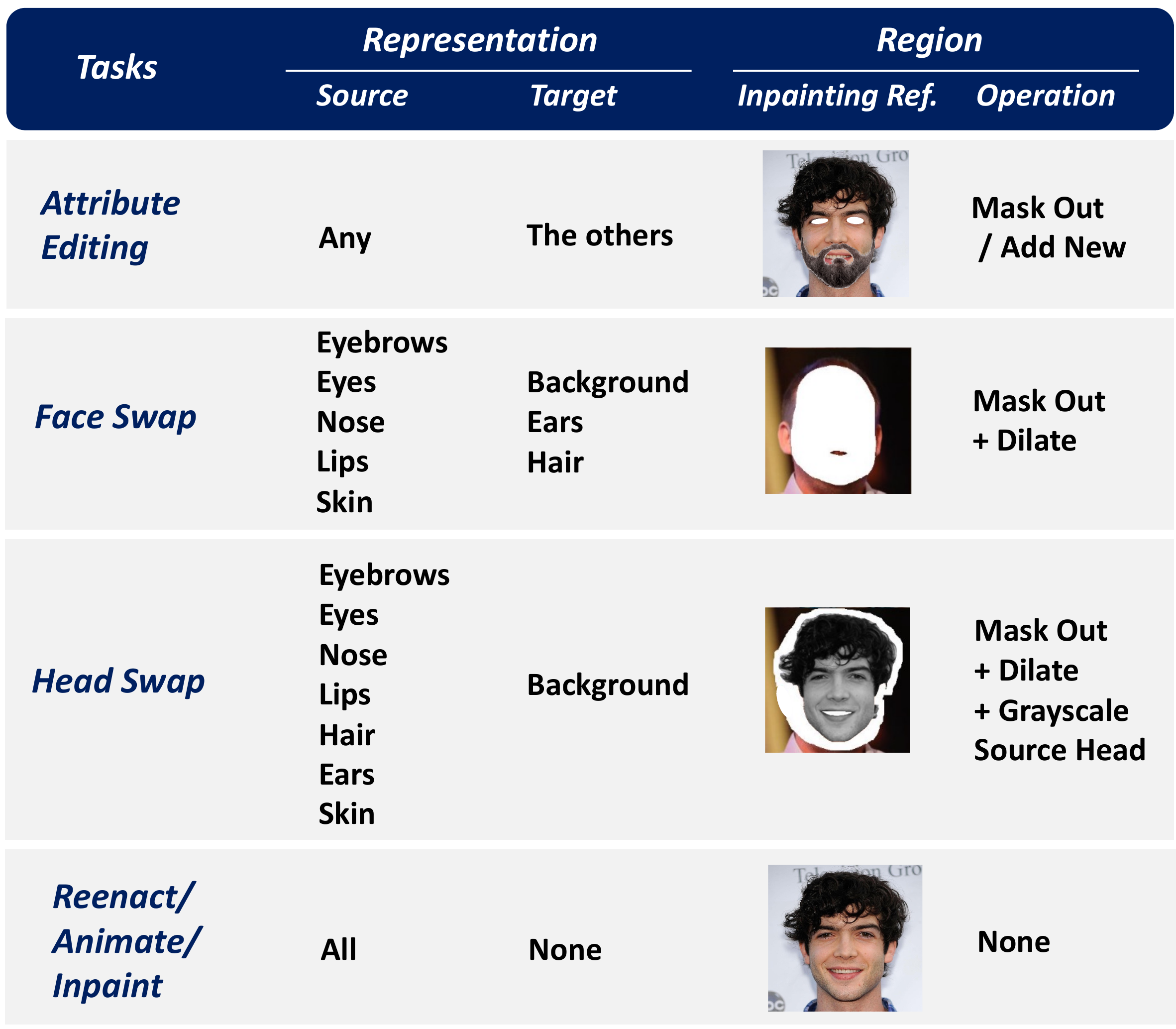}
    \caption{\textbf{Illustrations on task-specific representation and region assemblers}, showing omni-representation decomposing of popular facial tasks.
    The representation here indicates the region feature $\bm{f}^{R}$, encompassing facial texture, hair and background, as inherited from~\cref{fig:intro}. 
    However, with more detailed divisions, facial texture is further separated into eyebrows, eyes, nose, lips, ears, and skin.
    }
    \label{fig:taskdef}
    \vspace{-1.5em}
\end{figure}

\subsection{Training and Inference Details} 

\textbf{Generalist Model.} 
During training, both Task-specific Region and Representation Assemblers utilize the assembly method of head swapping. 
During testing, they perform according to the definitions of each task. 
This is because head swapping encompasses both reenactment and face swapping subtasks.
In a nutshell, our generalist single model is trained once and supports diverse facial editing tasks.

\noindent
\textbf{Specialized Models.} 
Other facial editing tasks have much lower requirements for region attribute disentanglement compared to head swapping task. 
To further improve the performance of subtasks, we finetune our model on these subtasks. In both training and testing, the Task-specific Region and Representation Assembler use the definition of the respective task.

\noindent
\textbf{Task-specific Representation Assembler.}
The representation combination methods for each task are defined in~\cref{fig:taskdef}.
For reenactment, all source region features are used. 
For face swapping, the eyebrows, eyes, nose, lips, and skin features of the source image are combined with other features of the target image. 
For head swapping, the eyebrows, eyes, nose, lips, hair, ears, and skin features of the source image are combined with other features of the target image. 

\noindent
\textbf{Task-specific Region Assembler.}
The region combination methods for each task are defined in~\cref{fig:taskdef}.
For face reenactment, the entire source image is used. 
For face swapping, the source face is recombined with the hair and background of the target. To avoid residual irrelevant information, the union of the source and target face areas is dilated. 
For head swapping, the grayscale source head is recombined with the target background, and the edges are cut out using dilation.

\section{Experiment}
\label{sec:experiment}

%
\noindent
\textbf{Dataset.}
We train~\name~on the CelebV~\cite{zhu2022celebv} dataset. 
For the face reenactment task, we evaluate on FFHQ~\cite{karras2019stylegan} and VoxCeleb1~\cite{nagrani2017voxceleb} test sets.
For face swapping tasks, we evaluate on FaceForensics++~\cite{rossler2019faceforensics++}(FF++). 
For head swapping tasks, we evaluate our model using FFHQ~\cite{karras2019stylegan} dataset. 
Additionally, we randomly collect images of well-known individuals from the Internet to demonstrate the qualitative results of each sub-task.

\noindent
\textbf{Metrics.}
We evaluate different methods from three perspectives: 
\textbf{\textit{1)} Motion.}  We assess the motion accuracy by calculating the average $L_2$ distance of pose, expression, and gaze embeddings between the generated and target faces. 
These three embeddings are derived through the respective estimator.
\textbf{\textit{2)} Identity.} We compute the cosine similarity of the identity feature between the generated and source faces. The identity feature is extracted by a face recognition model.
\textbf{\textit{3)} Image Quality.} We use the Fréchet Inception Distance (FID) to assess the quality of the generated faces.

\noindent
\textbf{Training Details.}
We start training from the StableDiffusion v1-5 model and OpenAI's clip-vit-large-patch14 vision model at a resolution of $256$.
For higher resolution of $512$ or $768$, we finetune on SD v$2.0$. 
As the head swapping task utilizes all framework components to encompass a comprehensive set of sub-capabilities, we designate the head-swapping model as our generalist model.
Training our generalist models entails 20k steps on $4$ V$100$ GPUs, at a constant learning rate of $1e-5$ and a batch size of $32$. 
Notably, for inpainting and animation tasks, no additional finetuning is needed. 
The generalist model inherently possesses robust inpainting capabilities. 
Moreover, during testing, animation tasks can be accomplished by directly loading community model weights.
For face reenactment and swapping tasks, we further finetune for $15$k and $5$k steps respectively with a subset of framework components.
To facilitate classifier-free guidance sampling, we train the model without conditions on $10$ of the instances. 

\subsection{Results of Popular Facial Tasks}

Our generalist model encapsulates the capabilities of all subtasks, liberating facial editing from fixed-structure appearance modifications in specific task, enabling dynamic facial edits, and enhancing the diversity of editing possibilities. 
However, the intricate disentanglement of representation and regions leads to a relative performance decrease in tasks that require less decoupling, \eg face reenactment and swapping. 
To address this, we fine-tune the generalist model on specific tasks to mitigate the performance drop caused by intricate disentanglement, enhancing metrics for these tasks.

\begin{figure}[t!]
    \centering
    \includegraphics[width=1\linewidth]{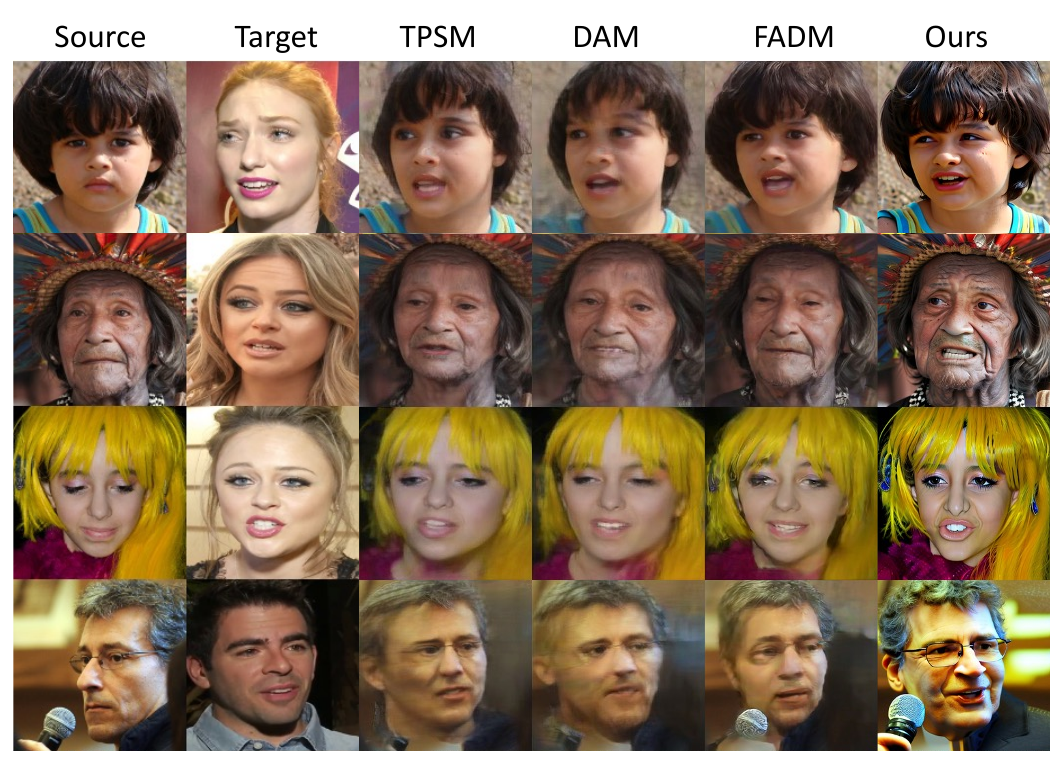}
    \vspace{-7mm}
    \caption{Qualitative comparison results on face reenactment.}
    \label{fig:reenact}
    \vspace{-1em}
\end{figure}

\begin{figure}[t!]
	\centering
	\includegraphics[width=1.0\linewidth]{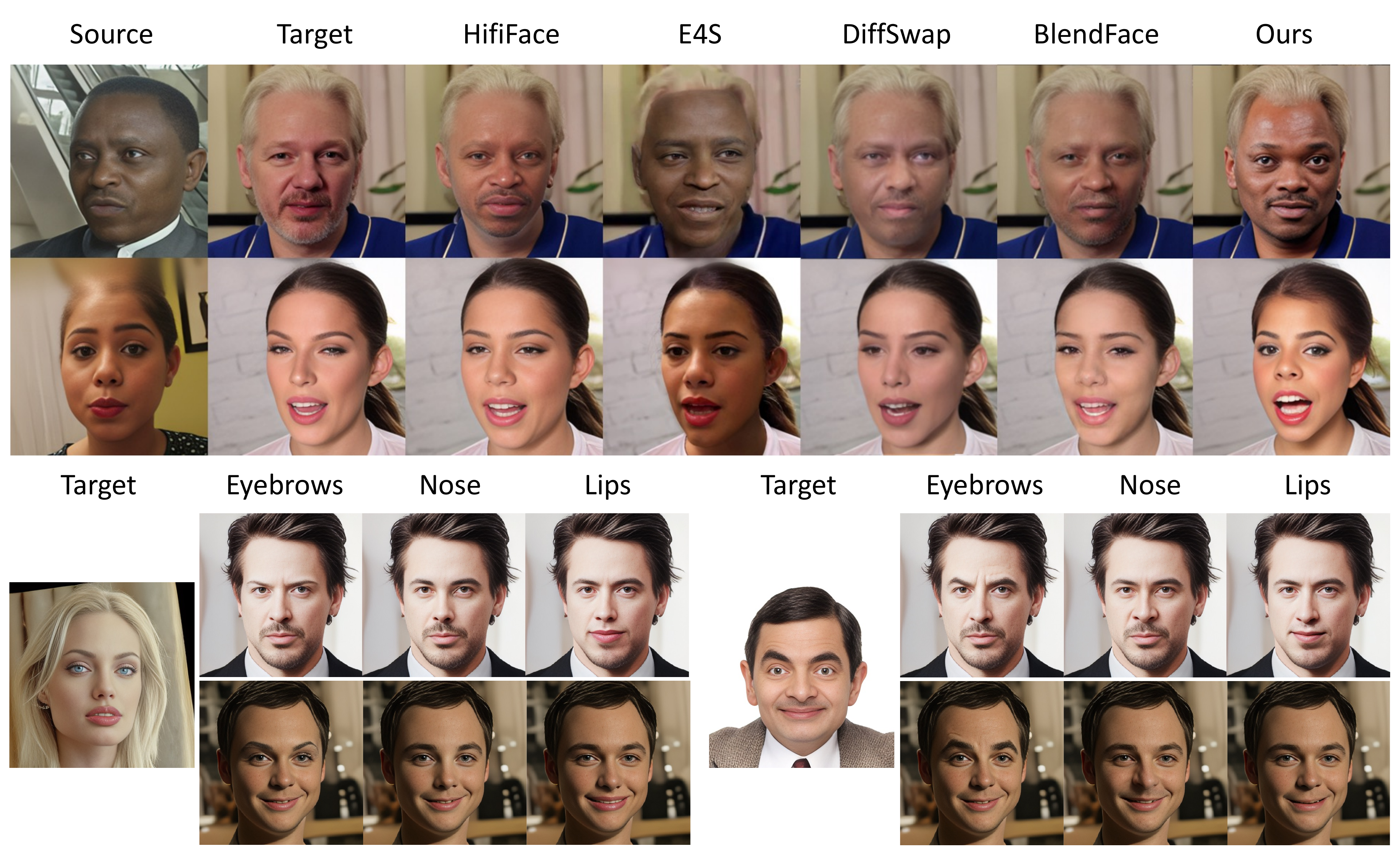}
	\vspace{-6mm}
	\caption{\textbf{Top:} Qualitative comparison results on face swapping. 
            \textbf{Bottom:} Controllable face swapping.}
	\label{fig:faceswap_c}
	\vspace{-1em}
\end{figure}

\begin{table}[]
\centering
\caption{Quantitative experiments on cross-identity face reenactment, using VoxCeleb test images to drive the FFHQ images. }
\label{tab:reenact}
\resizebox{\linewidth}{!}{%
\begin{tabular}{@{}lccccc@{}}
\toprule
                           & Exp Err.$\downarrow$ & Pose Err.$\downarrow$ & Gaze Err.$\downarrow$ & ID Simi.$\uparrow$ & FID$\downarrow$   \\ \midrule
CVPR'22 TPSM               & 6.10     & \underline{0.0535}     & 0.0900     & 0.5836    & 50.43 \\
CVPR'22 DAM                & 6.31     & 0.0626    & 0.0967    & 0.5534   & 54.13 \\
CVPR'23 FADM               & 6.71     & 0.0821     & 0.1242     & 0.6522    & \underline{42.22} \\
Ours-Generalist            & \underline{5.45}     & 0.0542     & \underline{0.0758}     & \underline{0.6612}    & 43.34 \\
Ours-Finetuned Specialized & \textbf{5.03}    & \textbf{0.0503}     & \textbf{0.0614}     & \textbf{0.6778}    & \textbf{35.67}  \\ \bottomrule
\end{tabular}
}
\end{table}

\begin{table}[]
\centering
\caption{Quantitative results for face swapping on FF++. }
\label{tab:faceswap}
\resizebox{\linewidth}{!}{%
\begin{tabular}{@{}lccccc@{}}
\toprule
                           & Exp Err.$\downarrow$ & Pose Err.$\downarrow$ & Gaze Err.$\downarrow$ & ID Simi.$\uparrow$ & FID$\downarrow$   \\ \midrule
IJCAI'21 HifiFace           & 5.50     & 0.0506     & \textbf{0.0650}    & 0.4971   & \textbf{21.88}   \\
CVPR'23 E4S                & \underline{5.23}     & \textbf{0.0497}     & 0.0791    & 0.4792   & 36.56 \\
Ours-Generalist            & 5.29     & 0.0503     & 0.0693    & \underline{0.5031}   & 44.32  \\
Ours-Finetuned Specialized & \textbf{5.14}     & \underline{0.0501}     & \underline{0.0674}    & \textbf{0.5088}   & \underline{36.24} \\ \bottomrule
\end{tabular}
}
\end{table}

\noindent\textbf{Face Reenactment.}
In~\cref{fig:reenact}, we compare~\name~with SoTA methods, including GAN-based TPSM~\cite{zhao2022tpsm}, DAM~\cite{tao2022dam}, and diffusion-based FADM~\cite{zeng2023fadm}. 
When handling unseen identities at the same resolution, our method consistently generates significantly superior results with richer texture details, \ie, teeth, hair, and accessories. 
Our approach maintains identity faithfully when source faces have different ethnicities, ages, extreme poses, and even occlusion.
\cref{tab:reenact} demonstrates our model delivers more precise motion control quantitatively.

\begin{figure}[t]
    \centering
    \includegraphics[width=1.0\linewidth]{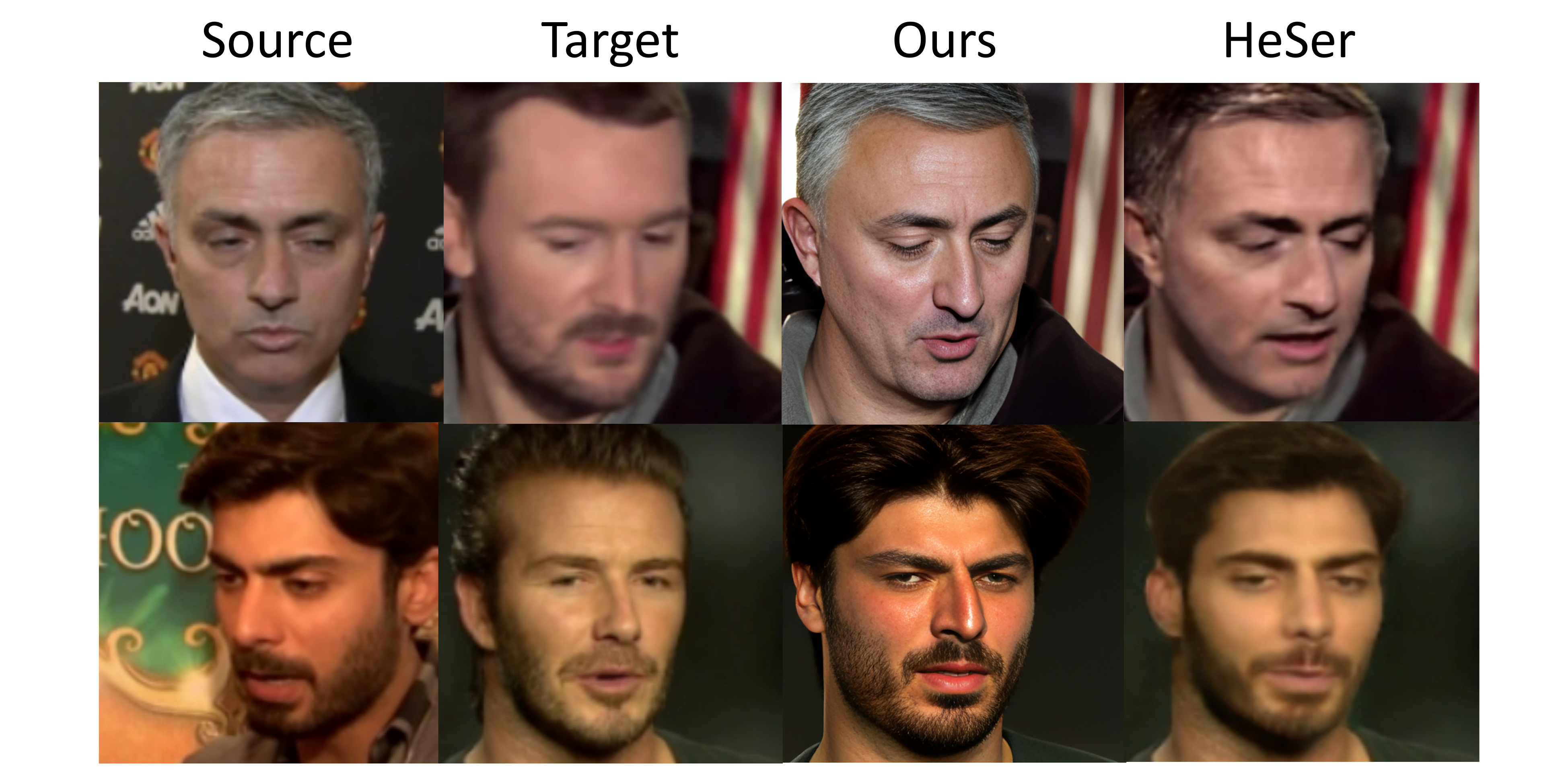}
    \vspace{-6mm}
    \caption{Qualitative comparison with HeSer on head swapping.}
    \label{fig:heser}
\end{figure}

\begin{figure}[t!]
    \centering
    \includegraphics[width=1.0\linewidth]{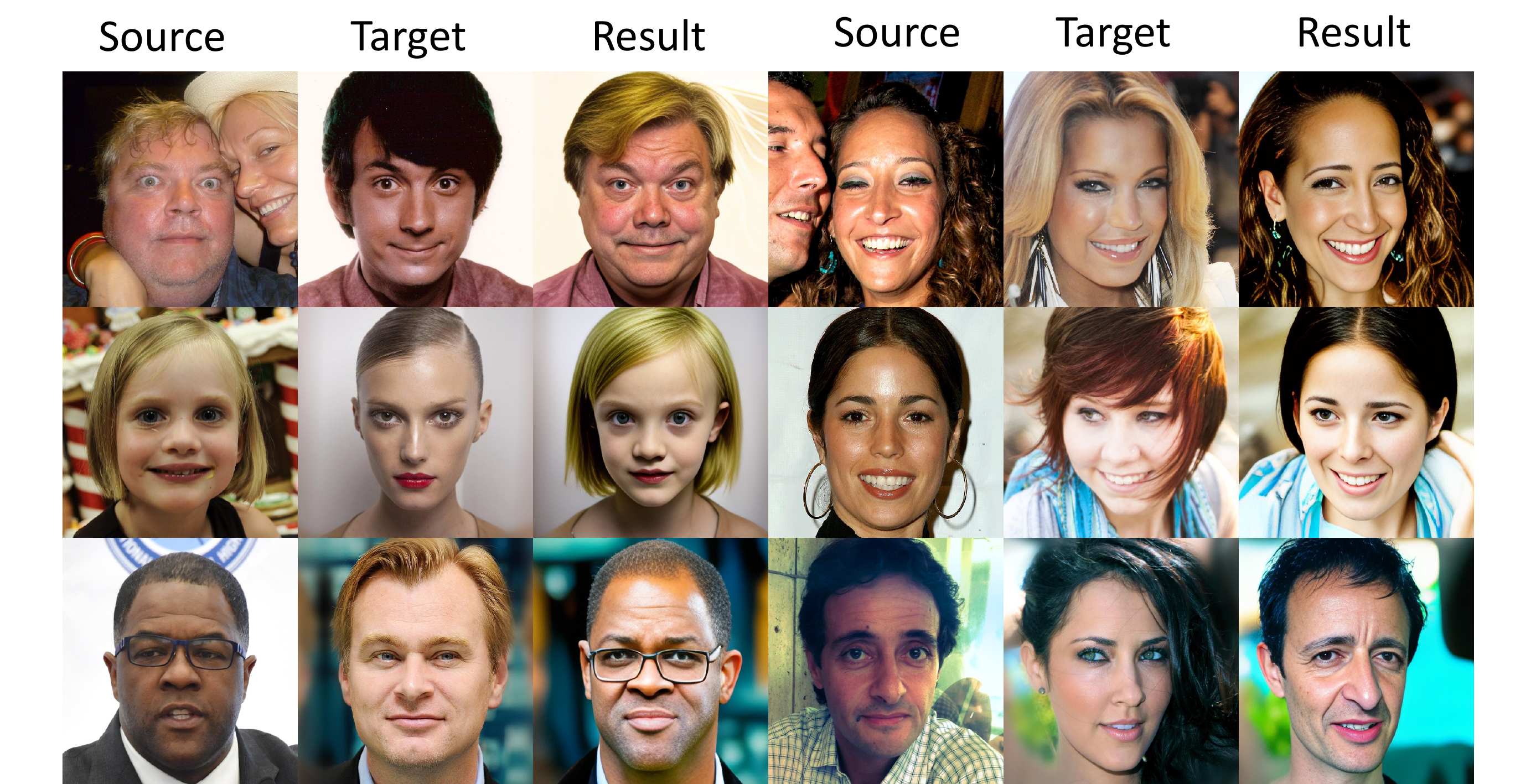}
    \vspace{-6mm}
    \caption{Qualitative results on head swapping.}
    \label{fig:swaphead}
\end{figure}

\begin{figure}[t!]
    \centering
    \includegraphics[width=1.0\linewidth]{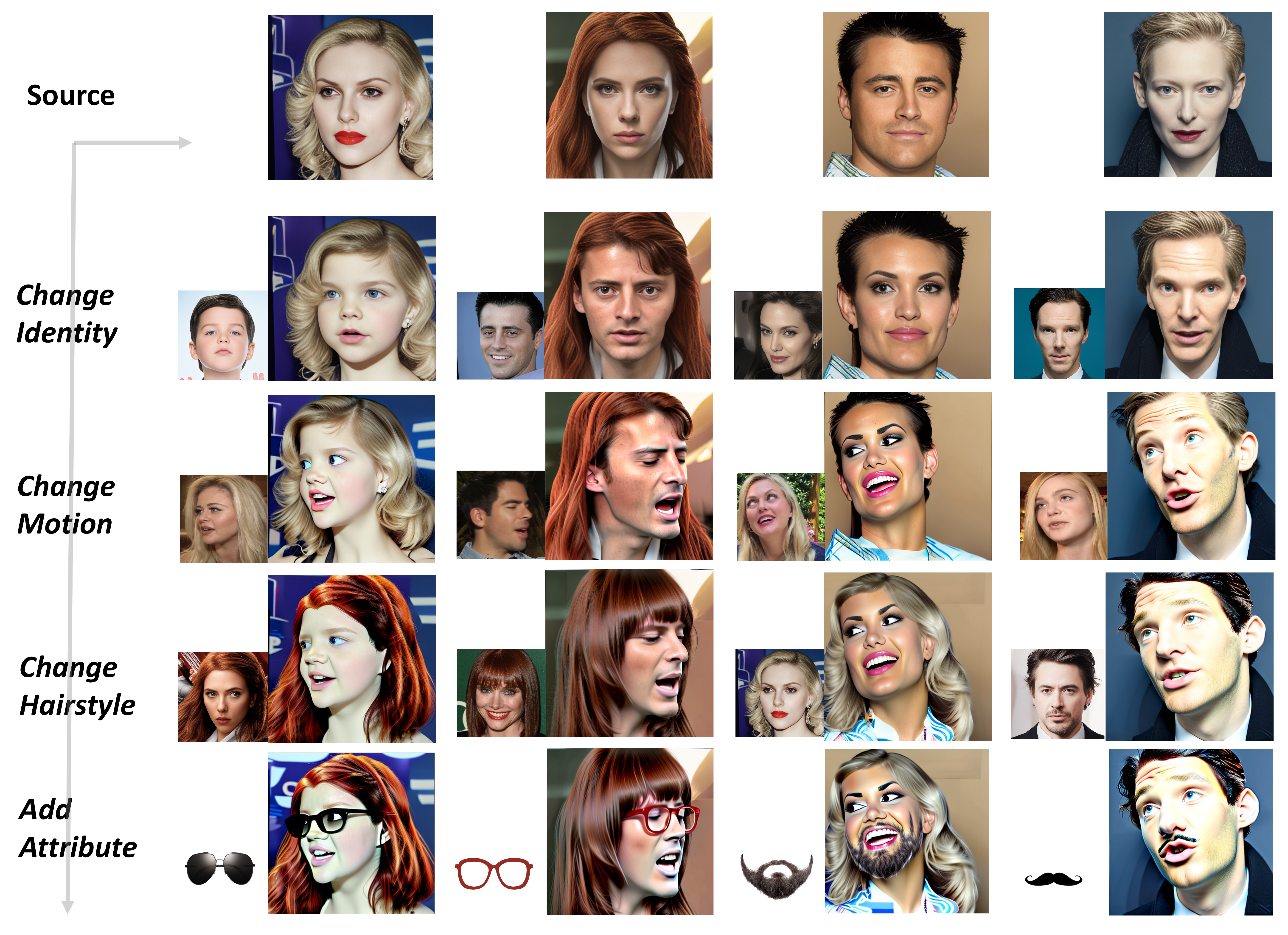}
     \vspace{-7mm}
    \caption{\textbf{Progressive Editing} using our generalist model.}
    \label{fig:progressive_edit}
\end{figure} 

\begin{figure}[t!]
	\centering
	\includegraphics[width=1.0\linewidth]{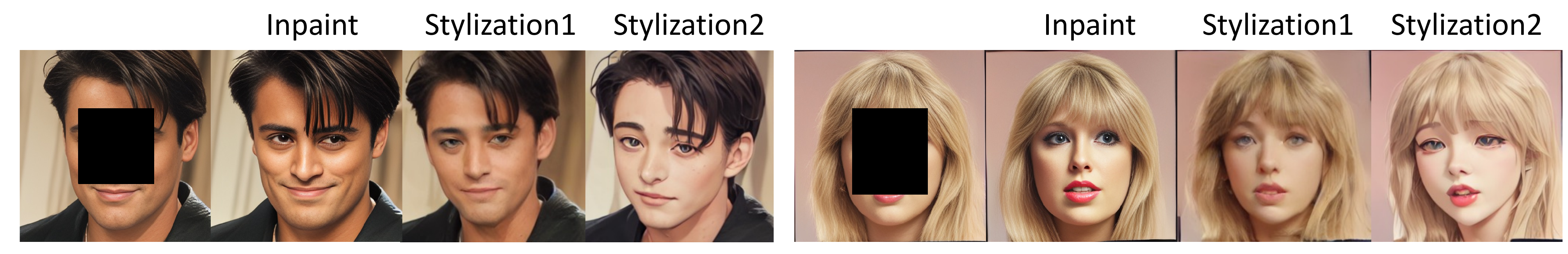}
	\vspace{-7mm}
	\caption{Extension to face inpainting and animation.}
	\label{fig:style_inpaint}
\end{figure}

\noindent\textbf{Face Swapping.}
\cref{fig:faceswap_c}-left shows a comparative analysis between~\name~and recent HifiFace~\cite{wang2021hififace} and E4S~\cite{liu2023e4s}.
HifiFace adopts a target-oriented strategy, emphasizing fidelity to the target in terms of facial color and texture. 
On the contrary, source-oriented E4S prioritizes adherence to the source characteristics. 
Our method strives to preserve the facial texture and certain skin color features from the source while maintaining harmony with the target environment.
Considering that E4S employs a face enhancement model to improve image resolution, to ensure fairness, we apply the same model to both HifiFace and our results.
\cref{fig:faceswap_c}-right shows the controllable attribute swapping results. 
By applying masked fusion during the inference sampling process, diffusion-based methods facilitate the selective swapping of a portion of the facial area, enabling the seamless integration of the substituted region with its surroundings.

Quantitatively, \name~exhibits competitive performance with SoTA methods in \cref{tab:faceswap}. 
E4S employs target face parsing masks to constrain the output image structure, ensuring strict alignment with the target. 
Consequently, it manifests a closer resemblance to the target in terms of both pose and expression. 
Our approach reduces structural constraints to enhance flexibility in motion control.



\noindent\textbf{Head Swapping.}
As HeSer~\cite{shu2022heser}, the recent SoTA, is not open-source, we compare using crops from the paper in~\cref{fig:heser}. 
Unlike target-oriented method HeSer, we prioritize source texture and skin color while harmonizing with the target.
HeSer uses multiple images of the source face to extract identity and perform a two-stage process by first reenacting the source face before conducting face swapping. 
In contrast, our one-shot-one-stage framework demonstrates comparable identity and motion consistency while achieving much higher image quality. 
Further, ~\cref{fig:swaphead} evaluates~\name~on datasets with \textit{more complex environment} beyond the VoxCeleb dataset used by HeSer, where lighting conditions are consistently dim.  
The results show that our~\name~accurately maintains skin color across various ethnicities and adapts to the target lighting conditions.

\noindent\textbf{Progressive Editing across Diverse Facial Tasks.}
\cref{fig:progressive_edit} illustrates the diverse facial editing capabilities of our generalist model, showcasing the progressive achievement of editing identity, motion, and semantic attributes.
Note that the arrangement and order of facial features may be arbitrary.
%
%
In contrast to previous methods limited by fixed structures, our approach supports flexible combination of different editing capabilities, enhancing the diversity of editing possibilities. 

\begin{figure*}[t!]
	\centering
	\includegraphics[width=1.0\textwidth]{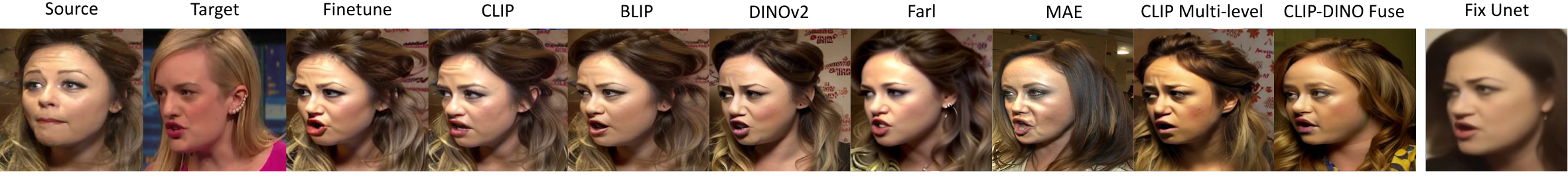}
	\vspace{-7mm}
	\caption{\textbf{Left:} Ablation of using different visual encoders. \textbf{Right:} Fixing U-net without FRC results in a failure to reconstruct texture. }
	\label{fig:visual_abla}
\end{figure*}

\begin{figure}[h]
    \centering
    \includegraphics[width=1.0\linewidth]{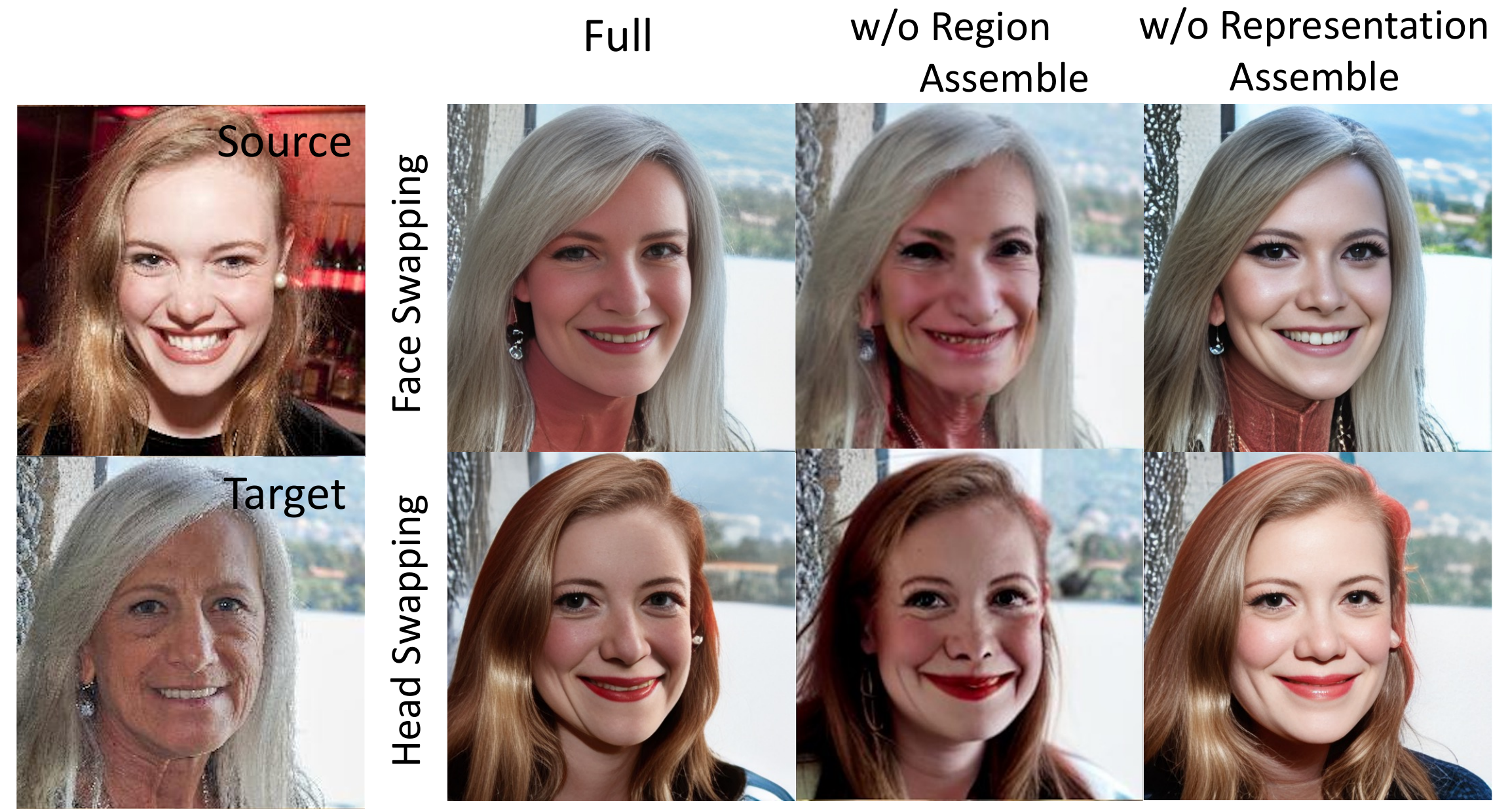}
    \vspace{-7mm}
    \caption{Qualitative comparison of our model under different ablative configurations.}
    \label{fig:ablation_facex}
    \vspace{-1em}
\end{figure}

\noindent
\textbf{Inpainting and Animation.}
Benefiting from our fine-tuning strategy, freezing the U-net weights during training and loading community personalized model weights during testing enables us to achieve stylization. 
\cref{fig:style_inpaint} showcases animated stylizations with watercolor and oil painting brushstrokes.
On the other hand, our method demonstrates a robust inpainting capability by retaining SD prior knowledge. This is evident in its ability to generate reasonable facial inpainting results, even when confronted with substantial facial voids.


\subsection{Ablation Study}
\label{se:ablation}

\noindent\textbf{Choice of Visual Encoders.}
We ablate different visual encoders in \cref{fig:visual_abla}, \ie, CLIP-based ViT~\cite{radford2021clip,vit}, DINOv2~\cite{oquab2023dinov2}, FARL~\cite{farl}, BLIP~\cite{blip}, and MAE~\cite{mae}, on face reenactment, because facial tasks may heavily rely on the representations from pre-trained models. 
We draw the following conclusions: 
\textbf{\textit{1)}} Finetuning visual encoders exhibits a significantly faster convergence than fixing them. Despite variations in convergence speed, different models of ViT ultimately yield closely aligned results. 
\textbf{\textit{2)}} Initialization via the weights of CLIP ViT demonstrates the fastest convergence during finetuning. 
The obtained results are also superior with fixed weights. 
This phenomenon might be attributed to the alignment between the visual branch of CLIP and the text branch of SD.
\textbf{\textit{3)}} Under fixed weights, the performance hierarchy is as follows: CLIP > DINOv2 = BLIP > FARL > MAE. 
Neither the fusion of multi-stage features from CLIP ViT nor a combination of features from CLIP and DINOv2 yields superior results.

\noindent\textbf{Task-specific Region Assembler.}
Due to the structural information loss caused by mask pooling in the Task-specific Region Assembler, removing this assembler results in the model lacking direct guidance from structural information. Hence, the model tends to generate ambiguous outcomes, which is demonstrated in \cref{fig:ablation_facex} and \cref{tab:ablation}. 

\noindent\textbf{Task-specific Representation Assembler.}
Task-specific Region Assembler can only provide structural guidance, and it requires the Task-specific Representation Assembler to supply local appearance information. 
If this information is lacking, it can lead to color bias in the generated results.

\noindent\textbf{Facial Representation Controller.}
When the U-net is frozen and FRC is removed, solely finetuning the FORS module may enable the model to capture coarse identity and motion. 
Thus, generating detailed textures becomes difficult as shown in \cref{fig:visual_abla}-right.

\subsection{Discussion on Efficiency}
\label{se:ablation}
As a diffusion-based method, our approach does not exhibit a computational advantage in terms of inference time when compared to GAN-based methods, including TPSM, DAM, and HifiFace. 
However, we distinguish ourselves by achieving a notable advantage in image quality. Specifically, in contrast to the face swapping method E4S, which requires pre-alignment using a reenactment model, our method achieves uniformity within a single model. Additionally, head swapping method HeSer necessitates fine-tuning on \textit{multiple} images of the source identity, whereas we accomplish identity preservation in a \textit{one-shot }manner.
Compared to other diffusion-based methods, FADM involves obtaining a coarse driving result using a previous reenactment model, followed by refinement using DDPM. 
In contrast, our method operates as a unified model. 
Regarding training costs, our model freezes the parameters of the SD Unet and only fine-tunes the additional introduced parameters. 
This leads to faster convergence compared to FADM, which trains from scratch. 

\begin{table}[]
\centering
\resizebox{0.45\textwidth}{!}{%
\begin{tabular}{@{}lcccc@{}}
\toprule
Configuratons               & SSIM$\uparrow$ & PSNR$\uparrow$& RMSE$\downarrow$& FID$\downarrow$\\ 
\midrule
w/o Region Assemble         & 0.6580         & 14.79         & 3.32            & 45.31        \\
w/o Representation Assemble & 0.7520         & 18.24         & 1.78            & 29.27        \\ 
Our Full Model              & 0.7960         & 19.15         & 1.31            & 27.95        \\
\bottomrule
\end{tabular}
}
\vspace{-3mm}
\caption{Quantitative comparison of our model under different ablative configurations. The reconstruction performance is measured.}
\label{tab:ablation}
\end{table}

\section{Conclusion and Future Works} \label{sec:conclusion}

In this paper, we propose a novel generalist~\name~to accomplish a variety of facial tasks by formulating a coherent facial representation for a wide range of facial editing tasks.
Specifically, we design a novel FORD to easily manipulate various facial details, and a FORS to first assemble unified facial representations and then effectively steer the SD-aware generation process by the designed FRC. 
Extensive experiments on various facial tasks demonstrate the unification, efficiency, and effectiveness of the proposed method.

\noindent\textbf{Limitations and Future Works.} As this paper aims to design a general facial editing model, it may be suboptimal on some metrics for certain tasks. 
In the future, we will further explore more effective methods, including investigating the integration of large language models or large vocabulary size settings~\cite{touvron2023llama,wu2023open} for task expansion.

\noindent\textbf{Social Impacts.} Generating synthetic faces increases the risk of image forgery abuse. In the future, it's necessary to develop forgery detection models in parallel to mitigate this risk.

{
    \small
    \bibliographystyle{ieeenat_fullname}
    \bibliography{main}
}


\end{document}